\documentclass[12pt]{article}
\usepackage{amsmath}
\usepackage{times}
\usepackage{graphicx}
\usepackage{color}
\usepackage{multirow}
\usepackage{rotating}
\usepackage{bbm}
\usepackage{latexsym}

\usepackage[style=apa,backend=biber,language=american]{biblatex}
\DeclareLanguageMapping{american}{american-apa}
\usepackage[modulo]{lineno}
\newcommand{\citep}{\parencite}
\newcommand{\citet}{\cite}

\newcommand{\captionfonts}{\normalsize}

\makeatletter  
\long\def\@makecaption#1#2{%
  \vskip\abovecaptionskip
  \sbox\@tempboxa{{\captionfonts #1: #2}}%
  \ifdim \wd\@tempboxa >\hsize
    {\captionfonts #1: #2\par}
  \else
    \hbox to\hsize{\hfil\box\@tempboxa\hfil}%
  \fi
  \vskip\belowcaptionskip}
\makeatother   
\begin{document}
\hspace{13.9cm}1

\ \vspace{5mm}\\

\ \\
{\bf \large Robustness to Transformations Across Categories: \\ Is Robustness Driven by Invariant Neural Representations?}\\ [0.5em]

 \noindent{Hojin Jang$^{1,\star}$  \; \; \;  \; \; \; Syed Suleman Abbas Zaidi$^{1,2,\star}$ \\
 Xavier Boix$^{1,3,4\star, \bullet}$
 \; \; \;  \; \; \; Neeraj Prasad$^{1,\star}$\\ %[-2em]
 Sharon Gilad-Gutnick$^1$ 
 \; \; \;  \; \; \;  \;   { Shlomit Ben-Ami$^1$} 
  \; \; \;  \; \;  \;  { Pawan Sinha$^{1}$ } \\ 
  
  \noindent $^1$ Department of Brain and Cognitive Sciences, Massachusetts Institute of Technology (USA) \\
  $^2 $ Department of Electrical and Computer Engineering, Technical University of Munich (Germany) \\
 $^3$ Center for Brains, Minds and Machines (USA) \\ 
  $^4$ Fujitsu Research of America (USA) \\ 
 $^\star$ \emph{Shared first authorship.} \\
  $^\bullet$ \emph{Corresponding author:} \texttt{xboix@fujitsu.com}\\
}
  
 %{ Keywords:} Generalization Beyond the Training Distribution, Individual Neuron Invariance, Robustness to Transformations

% Attempt to make hyperref and algorithmic work together better:
\newcommand{\theHalgorithm}{\arabic{algorithm}}

\newcommand*{\eg}{\emph{e.g.~}}
\newcommand*{\ie}{\emph{i.e.~}}
\newcommand*{\etal}{\emph{et al.~}}
\newcommand*{\etc}{\emph{etc.}}

% The \author macro works with any number of authors. There are two commands
% used to separate the names and addresses of multiple authors: \And and \AND.
%
% Using \And between authors leaves it to LaTeX to determine where to break the
% lines. Using \AND forces a line break at that point. So, if LaTeX puts 3 of 4
% authors names on the first line, and the last on the second line, try using
% \AND instead of \And before the third author name.

\begin{abstract}
Deep Convolutional Neural Networks (DCNNs) have demonstrated impressive robustness to recognize objects under transformations (\eg blur or noise) when these transformations are included in the training set. A hypothesis to explain such robustness is that DCNNs develop invariant neural representations that remain unaltered when the image is transformed. However, to what extent this hypothesis holds true is an outstanding question, as robustness to transformations could be achieved with properties different from invariance,~\eg parts of the network could be specialized to recognize either transformed or non-transformed images. This paper investigates the conditions under which invariant neural representations emerge by leveraging that they facilitate robustness to transformations beyond the training distribution. Concretely, we analyze a training paradigm in which only some object categories are seen transformed during training and evaluate whether the DCNN is robust to transformations across categories not seen transformed. Our results with state-of-the-art DCNNs indicate that invariant neural representations do not always drive robustness to transformations, as networks show robustness for categories seen transformed during training even in the absence of invariant neural representations. Invariance only emerges as the number of transformed categories in the training set is increased. This phenomenon is much more prominent with local transformations such as blurring and high-pass filtering than geometric transformations such as rotation and thinning, which entail changes in the spatial arrangement of the object. Our results contribute to a better understanding of invariant neural representations in deep learning and the conditions under which it spontaneously emerges.

\end{abstract}
\clearpage
\section{Introduction}

%PARAGRAPH 1 - Phenomenon under analys: robustness to transformations with data augmentation. Cite Geiros and works on data augmentation. 
A widely-known strategy to gain robustness to image and object transformations in Deep Convolutional Neural Networks (DCNNs) is to include the transformation in the training set~(\cite{wang2017effectiveness}, \cite{schmidt2018adversarially}, \cite{NIPS2018_7982}, \cite{geirhos2018imagenet}). \cite{NIPS2018_7982} recently demonstrated that this strategy is successful in achieving robustness to local transformations, such as blur and noise, and also outperforms human-level recognition accuracy in transformed images. In a subsequent paper, \cite{geirhos2018imagenet} show that including some transformations in the training set  that emphasize the shape of the object rather than the texture leads to robustness to transformations beyond the ones seen during training.

%This suggests that generalization beincluding transformations in the training set could be a.

%Architectural choices such as multiscale architectures and larger feature-aggregating models alleviate the DCNN fragility to transformations~\cite{hendrycks2018benchmarking}.

%PARAGRAPH 2 - The mechanisms of the robustness are poorly understood. There is one main hypothesis: invariance, 
Despite these recent results demonstrating effective ways of achieving robustness to transformations, the field still lacks a good understanding of the underlying neural mechanisms that enable such robustness. Decades of research at the intersection of Neuroscience and Computer Vision has led to the hypothesis that neural networks may develop invariant neural representations to achieve robustness to transformations~(\cite{riesenhuber1998just}, \cite{quiroga2005invariant}, \cite{goodfellow2009measuring}, \cite{achille2018emergence}, \cite{poggio2016visual}),~\ie the internal neural representations may remain mostly unaltered when the image is transformed~\footnote{We use the term \emph{robustness to a transformation} (or simply \emph{robustness}) to denote that DCNN classification accuracy does not deteriorate in transformed images, whereas we use the term \emph{invariance to a transformation} (or \emph{invariance}) to indicate that the internal neural representation of the DCNN are not altered by the transformation.  }. Yet, to what extent this hypothesis holds true is an outstanding question. It is possible that to achieve robustness,  neural representations develop properties different from invariance  when transformations are included in the training set,~\eg parts of the network may specialize in recognizing either transformed or non-transformed images.

%In this paper, we investigate to what extent and under what conditions invariant representations emerge in DCNNs trained with transformations. 
%To do so, we focus on a fundamental property of invariant representations, which is that they enable robustness to object categories that the network has not seen with transformation during training~\citep{poggio2016visual}. We denote this form of robustness as \emph{across-category robustness}, while we use the term \emph{within-category robustness} to designate robustness to categories that are seen transformed during training. 
%We evaluate  across-category robustness by analyzing DCNNs trained such that the transformations in the training set are restricted to a subset of categories. This evaluation serves to assess whether invariance is epiphenomenal or drives robustness beyond the categories seen transformed.

{\color{black}
To understand the emergence of invariant neural representations in DCNNs trained with transformations, an important question to address is to what extent and under what conditions such representations arise. 
Measuring the degree of invariance in neural representations can offer some hints but falls short of providing a comprehensive answer. A high degree of neural representations does not guarantee robustness to transformations, as even a small change in the representations can potentially result in poor robustness (\eg adversarial examples). Only when the neural representations exhibit perfect invariance to image degradations we can ensure robustness to transformations.  In this paper, we aim to study the potential relationship between invariance and robustness to transformations by examining the impact of invariant neural representations on robustness beyond the training distribution. }
Invariant neural representations imply that if the network correctly classifies images from a category, it will correctly classify transformed versions of the same image as well. This fundamental property of invariant neural representations enables robustness to object categories that the network has not seen with transformations during training~\citep{poggio2016visual}. Thus, we analyze a training paradigm in which only some object categories are seen transformed  during training, and evaluate robustness to transformations for object categories that the network has not seen transformed. We denote this form of robustness as \emph{across-category robustness}, while we use the term \emph{within-category robustness} to designate robustness to categories that are seen transformed during training. 
In this way,  we assess the emergence of invariant neural representations by evaluating \emph{across-category robustness}. To strengthen the evidence obtained with such evaluation, we also define a metric to directly measure the degree of invariance in the neural representations. This metric, together with the  \emph{across-category robustness} evaluation, provides the relationship between invariant neural representations and robustness to transformations.

%Measuring the degree of invariance in the neural activity provides  some hints towards an answer, but not a complete one, as a high degree of invariance does not necessarily mean that it relates to robustness,~\ie invariance could be an epiphenomenon with respect to robustness. We provide some reassurance that invariance relates to robustness by analysing a property of the network implied by invariance. whether the network has the following fundamental property  

%Furthermore, humans have across-category robust recognition, an ability that is used in everyday life; when we see an object for the first time, we can recognize it transformed with new illumination conditions, viewpoints,~\etc without having previously seen the object transformed in these ways. 

  A series of experiments with state-of-the-art DCNNs in FaceScrub~\citep{ng2014data} and ImageNet~\citep{deng2009imagenet} reveal that robustness is not always related to invariant neural representations, as networks show robustness to categories seen transformed during training even in the absence of invariant neural representations.   Invariance only emerges as the number of transformed categories in the training set is increased. {\color{black} Moreover, our observations indicate that different categories of transformations exhibit varying degrees of invariance, with local transformations (\eg blurring and high-pass filtering) exhibiting an increase in invariance roughly two times greater than that of geometric transformations (\eg rotations and thinning). } Our analysis uncovers new evidence that invariance emerges at the individual neural level and demonstrates that it facilitates robustness to categories that the network has not seen transformed during training. 
  
  These results are the first milestone to understand the neural representations that contribute to robustness to transformations. Furthermore, answering what representations emerge in the network could guide research to develop more robust models, and investigating \emph{across-category robustness} could help to reduce the amount of training examples, as not all categories need to be seen transformed to achieve robustness.

\begin{figure*}[t!]
\centering
\scriptsize
\includegraphics[width=\textwidth]{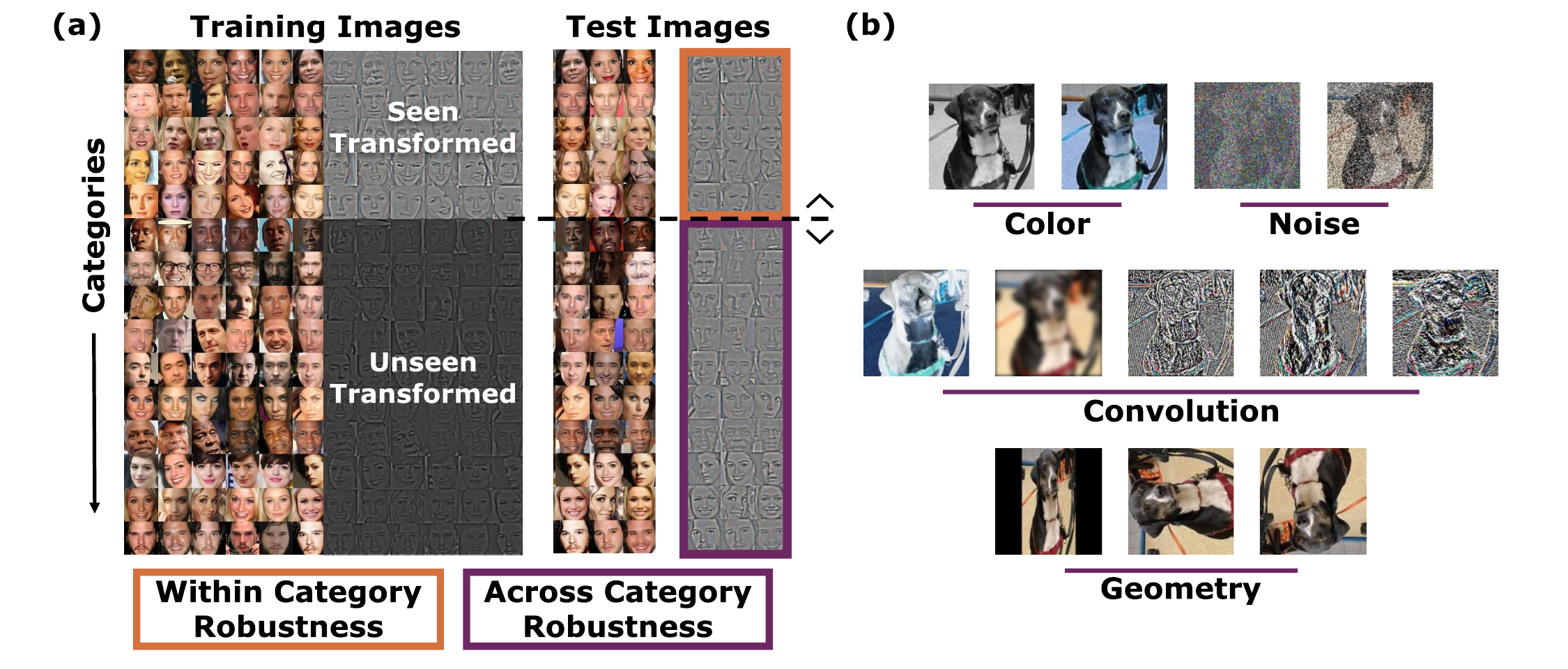}
\caption{\emph{Experimental Paradigm and Image Transformations} (a) The DCNN is trained with the transformations in the training set restricted to the \emph{seen-transformed} categories. The accuracy for transformed images is evaluated in categories that are either \emph{seen} or \emph{unseen} transformed.  (b) We analyze $12$ image transformations from four families: color, convolutional, noise and geometric.  }
\label{fig:paradigm}
\vspace{1cm}
\end{figure*}

\section{Signatures of Invariant Neural Representations}

\label{sec:signatures}

%Reproducing results - set up phenomenon under analysis

In this section, we introduce our methodology for investigating to what extent invariant neural representations contribute to robustness to transformations. For simplicity, we consider DCNNs trained on a set of images with only one transformation, as well as a set of non-transformed images.  This  yields robustness to the trained transformation but is unlikely to result in robustness to other transformations that the DCNN was not trained on~\citep{NIPS2018_7982}.

A neural network is completely invariant when the neural representations in the DCNN are identical for the transformed and non-transformed images. While this is ideal, it is important to note that it is not a pre-requisite for robustness: in practice, a certain amount of invariance may emerge, even without attaining identical neural activity, and result in robustness, {\color{black} that is, perfect invariance may not be needed in order to achieve robustness. The opposite may be true as well, as a certain amount of invariance may be unrelated to robustness,~\ie a high but imperfect degree of invariance may be epiphenomenal. 
To confirm the relationship between invariance and robustness, we evaluate across-category robustness in conjunction with the degree of invariance, which can offer strong evidence for the role of invariance in achieving robustness.} In the following, we first introduce the paradigm to evaluate across-category robustness, and then, the measure of invariance to transformations.

%This adds a difficulty in the analysis of invariance because it is difficult to determine whether small changes in the neural activity relate to robustness.  %Adversarial perturbations are one such example~\cite{szegedy2013intriguing}.

\subsection{Paradigm to Evaluate the Across-Category Robustness} 
\label{sec:paradigm}

To test across-category robustness we use the procedure depicted in Fig.~\ref{fig:paradigm}. Images of the subset of categories denoted as \emph{seen-transformed} appear both transformed and non-transformed in the training set. The remainder of the categories, denoted \emph{unseen-transformed}, appear in the training set only in their non-transformed version. Note that the sum of the number of categories in each set is equal to the total number of categories in the dataset. The network is trained with both \emph{seen-} and \emph{unseen-transformed} categories by uniformly randomizing the categories and images during each training step. When an image of the \emph{seen-transformed} set is used, it is also randomly chosen with equal probability to appear as transformed or not transformed.

The network's task is to predict the category of the input image.  We evaluate the accuracy of the network on transformed test images for the \emph{seen-} and \emph{unseen-transformed} categories separately, which yields the within- and across-category accuracy, respectively. We train multiple networks with different number of  \emph{seen-transformed} categories during training.
In all the experiments, the network predicts a category among all categories in the dataset,~\ie the possible output choices of the network remain the same when evaluating within- and across-category accuracy. 
 Thus, robustness among all the networks can be compared without bias because the only change between them is the number of \emph{seen-transformed} categories during training. {\color{black} Alternative paradigms are also available in addition to our proposed one, and more details can be found in Appendix~\ref{app:alterparadigms}.}

\subsection{Invariance to Transformations at the  Individual Neural Level}
\label{sec:invariance}

Understanding DCNNs at the individual neural level has led to promising results that provide granular and simple explanations of  the underlying mechanisms behind DCNN generalization~(\cite{zeiler2014visualizing}, \cite{zhou2018revisiting}, \cite{olah2018building}).
Individual neurons are commonly interpreted as feature detectors that are tuned to image features. A neuron is active,~\ie the neuron's output value is high, when the feature is present in the image, otherwise the neuron is not active and the output value of the neuron is low. Neurons are tuned to features that are not necessarily interpretable by humans~\citep{bau2017network}. We analyze the invariance of a neuron by evaluating the change of its activity when the image is transformed. For example, a neuron may be tuned to detect a dog nose because it is only active in the presence of a dog nose in the image. If the neuron is invariant, it will continue being active when the dog nose is transformed.

Invariance in neurons that are active may help to achieve robustness to transformations, but this is not the case for inactive neurons. If a neuron is inactive in both the transformed and the non-transformed images, the activity is invariant to the transformation. Yet, this is because the feature that the neuron is tuned to is not present in the image, which does not contribute to the network's  classification accuracy. Also, neurons whose activity is constant across all images,~\ie neurons that are not tuned to any feature, are invariant and do not contribute to network's accuracy. Thus, invariance is helpful to achieve robustness to transformations only when  neurons that are activated by an image are also invariant to the transformation.

Following these intuitions, \cite{goodfellow2009measuring} introduced a measure of invariance that analyzes the  activity in images that generate the top-$1\%$ of the   neuron's output value. These images are transformed and invariance is evaluated by assessing whether the activity remains in the top-$1\%$ or not. This metric is effective to discard the cases of invariance that are not related to robustness. Yet, it is unclear to what degree the activity generated in a neuron by the rest of the $99\%$ of images that are not analyzed relates to robustness. In the following, we introduce a procedure that controls that the majority of the neural activity related to robustness is taken into account by the invariance measurement. We first introduce the procedure to determine when a neuron is active, and then, the measure of invariance of active neurons.

\noindent {\bf Active Neurons.} Let $\mathcal{X}$ be the images of the test set and let $x\in \mathcal{X}$ be one of its images. We refer to the transformed version of the image as $T(x)$. For the sake of notation simplicity, we consider  $T(x)$ to be the transformation that is included in the training set without specifying the type transformation. Let $\hat{f}_k(x)$ be the activity of the neuron indexed by $k$ when the network's input is  $x\in\mathcal{X}$. Note that $\hat{f}_k(x)$ can be multiplied by a factor without affecting the performance of the network, if the post-synaptic weights are divided by that factor. Thus, {\color{black}without the post-synaptic weights it is not possible to determine the relative importance of each neuron. Thus, we consider } the order of magnitude of $\hat{f}_k(x)$ not relevant and we discard it by normalizing $\hat{f}_k(x)$. We define $f_k(x)$ as the normalized activity of neuron $k$, such that $f_k(x)$ is equal to $1$ for the image that generates the maximum activity among transformed and not-transformed images. Note that the minimum value that $f_k(x)$ can take is $0$ because the neurons use ReLU activation~\citep{krizhevsky2012imagenet}.

We define the neuronal tuning of a neuron as the set of images that generate an activity above a threshold. Let $\tau$ be such a threshold, and let $\mathcal{A}_k$ be the set of images that generate an output value higher than $\tau$. Thus, $\mathcal{A}_k$ represents the neuronal tuning of neuron $k$ as it contains the set of images that yield the neuron active. $\mathcal{A}_k$ is the following subset of $\mathcal{X}$: $\mathcal{A}_k = \left\{x\in\mathcal{X} \; | \;\left( f_k(x) >\tau \right) \vee \left( f_k(T(x)) >\tau \right)  \right\}$, in which  $\vee$ is the ``logic or'' operator. Note that the activity generated by both transformed and non-transformed images is considered.

In order to determine the value of $\tau$, we need to take into account a compromise between the following two factors. On one hand, the value of $\tau$ should be as high as possible, such that the neuronal tuning is specific and invariance due to lack of activity is discarded. On the other hand, $\tau$ should be made as low as possible in order to include in $\mathcal{A}_k$ all the activity that is relevant for the network's robustness. To do so, we validate that the network's robustness is not affected by ablating all neurons except the active, given an image. Also, we validate that the network's accuracy is at chance when all the active neurons are ablated. These tests provide reassurance that the active neurons drives the network's robustness. In the experiments section, we show that neurons are active only on a small set of images and their activity is almost the only responsible for the network's robustness.

\noindent {\bf Measure of Invariance in Active Neurons.}
In order to measure the invariance of a neuron, only the images in the set $\mathcal{A}_k$ will be used. This is because, as previously discussed, invariance of neurons that are not active for a given image are not related to robustness. We define $h_k(x)$ as the invariance measure of neuron $k$ given an image $x\in \mathcal{X}$. We use the following expression based on the normalized difference between the neurons' activity of the transformed and non-transformed images:
\begin{equation}
h_k(x) = 1 - \left| \frac{f_k(T(x))  -  f_k(x)}{f_k(T(x))  + f_k(x)}    \right|.
\label{eq:H}
\end{equation}
We can see by analyzing Eq.~\eqref{eq:H} that $h_k(x)$ is equal to $1$ when the neuron is perfectly invariant,~\ie $f_k(T(x)) =  f_k(x)$, and is equal to $0$  when the neuron's activity varies maximally, such that $f_k(T(x))=0$ and $f_k(x)=1$ and vice versa. This invariance measure takes values between $0$ and $1$ depending on  how much the activity of $x$ and $T(x)$ varies. We define $I_k$ as the invariance coefficient of neuron $k$. $I_k$ is obtained by averaging  $h_k(x)$ over the set of images that the neuron is active, $\mathcal{A}_k$,~\ie $I_k = \frac{1}{\#\mathcal{A}_k}\sum_{x\in \mathcal{A}_k} h_k(x)$, in which  $\#\mathcal{A}_k$ is the cardinality of the set $\mathcal{A}_k$ (the number of images that neuron $k$ is active for). 
In the experiments, we analyze the distribution of  $I_k$ among neurons in a layer.

\section{Results}

We evaluate robustness to $12$ image transformations shown in Fig.~\ref{fig:paradigm} (see Appendix~\ref{app:transformations} for details). These are grouped in four transformation types: color (gray-scale and color rotation), geometric (thinning and rotation), convolutional (blur, high-pass and horizontal, vertical filtering and contrast inversion) and noise (white and salt \&  pepper). Our choice of transformations was motivated by studies showing that DCNNs are robust to them when they are included in the training set~\citep{NIPS2018_7982}.

To ensure that our results are not dataset- or architecture-dependent, we evaluate AlexNet~\citep{krizhevsky2012imagenet} in the FaceScrub dataset~\citep{ng2014data} and ResNet-18~\citep{he2016deep} in the ImageNet dataset~\citep{deng2009imagenet}. In  ImageNet  we only evaluate four transformations (blur, high-pass filtering, white noise and salt \&  pepper noise)  as some other transformations are already present in the training set and can not be excluded (note that in ImageNet objects appear at different orientations, sizes, contrasts and colors). We use the loss function, regularizers, optimizer and other hyperparameter choices reported in the literature, except for the learning rates that are tuned for each experiment (see Appendix~\ref{app:dataset} for more details).

\subsection{Results on Across-Category Robustness}

\begin{figure*}[t!]
\centering
\scriptsize
\includegraphics[width=\textwidth]{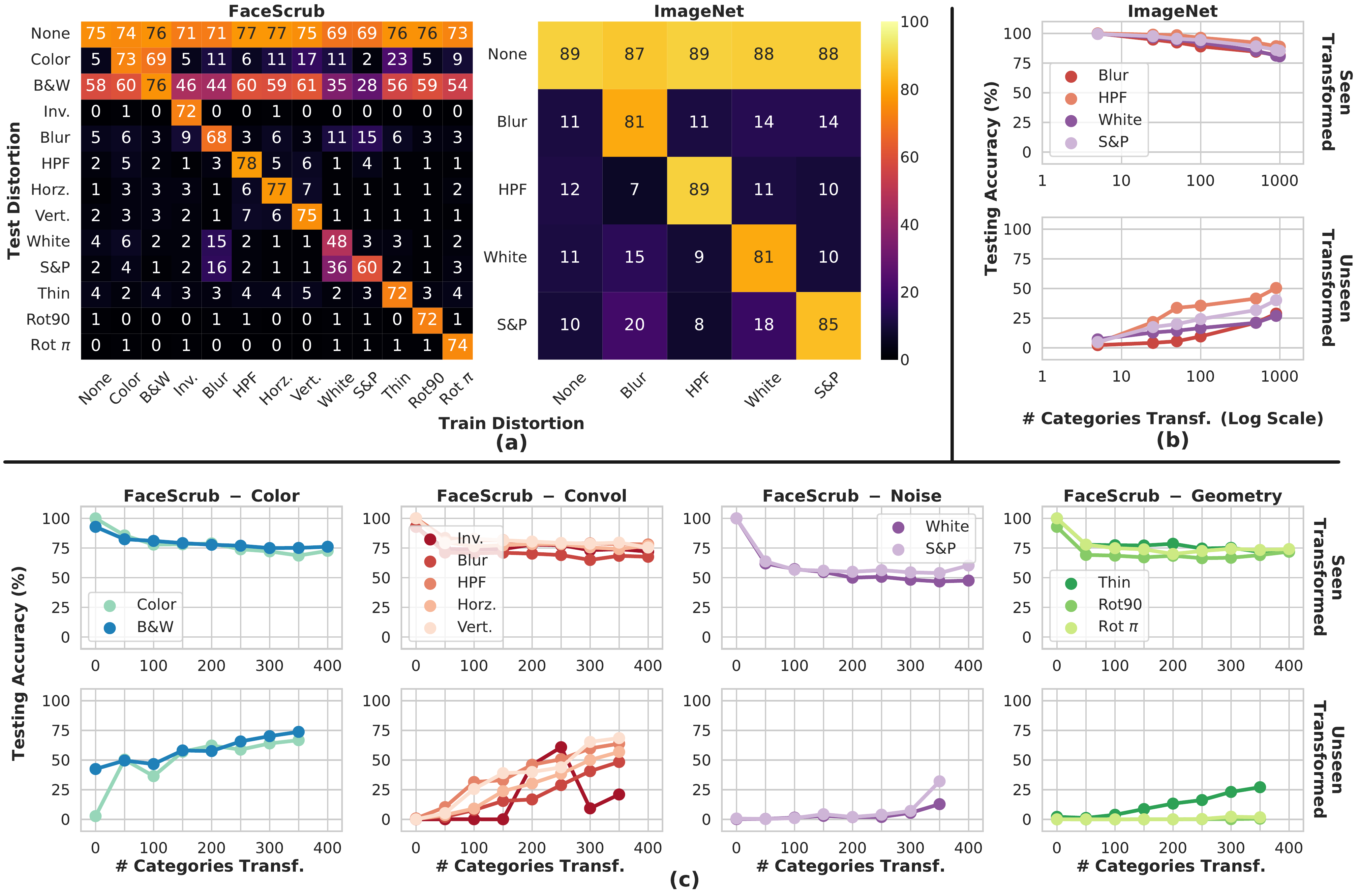}
\caption{\emph{Within- and Across-Category Accuracy in FaceScrub and ImageNet.} (a) Within-category accuracy when all object categories are seen transformed for FaceScrub (left) and ImageNet (right). (b) Within-category accuracy (top) and accros-category accuracy (bottom) in ImageNet, for different number of transformed categories in the training set. (c) Same as (b) for FaceScrub, each family of transformations is displayed separately.  }
\label{fig:fig2}
\vspace{1cm}
\end{figure*}

In the following, we report the accuracy of the network using the paradigm introduced in section~\ref{sec:paradigm}. For both FaceScrub and ImageNet, we report top-$1$ accuracy ($\%$).

\noindent {\bf Reproducing Previous Results on Robustness to Transformations.}  
Fig.~\ref{fig:fig2}a depicts the accuracy of networks trained with each transformation and tested on all transformations. As expected, on non-transformed images, the networks perform similarly to what is reported in the literature~(\cite{vogelsang2018potential}, \cite{he2016deep}) independently of  which transformations being included in the training set. On transformed images, networks are robust to the transformations included in the training but  not robust to other transformations, a finding consistent with~\cite{NIPS2018_7982}.  Note that color transformations are an exception, as  all networks have some degree of robustness to them. This may be because the color transformations are in part already included in the dataset as a consequence of different illumination conditions. Additionally, the network trained with blur transformation is slightly robust to the noise transformations, likely because training with blur yields large receptive fields in the first layer that  may help to filter out the noise ---a finding consistent with previous studies~\citep{vogelsang2018potential}.

%Geirhos~\etal~\cite{} evaluated the robustness when all object categories are transformed,~\ie the  within-category robustness.

\noindent {\bf Across-Category Robustness Improves as the Number of \emph{Seen-transformed} Categories is Increased.} 
In Fig.~\ref{fig:fig2}b and c, the within- and across-category accuracy are evaluated for different number of \emph{seen-transformed} categories. 
The within-category accuracy (in the top of the figure) and the across-category accuracy (in the bottom) are reported. We can observe that the across-category accuracy increases as more categories are seen transformed. This indicates that invariant neural representations may have emerged in the network (we can not confirm this without analyzing the neural activity, as across-category robustness is a consequence of invariance but not the reverse). Also, observe that when the network is trained with few \emph{seen-transformed} categories, the robustness for these categories can not be driven by invariance because the network is not across-category robust. Understanding the mechanisms for robustness when there is lack of invariance is an interesting open question that will be analyzed in future works. 

Note also that the within-category accuracy decreases as more  categories are seen transformed, and it reaches the values previously reported in Fig.~\ref{fig:fig2}a when all categories are seen transformed. Observe that the within-category accuracy is close to $100\%$ when the number of \emph{seen-transformed} categories is close to $1$.  This suggests that the network uses the  transformation as a feature to identify the few \emph{seen-transformed} categories. This can be validated with the confusion matrices between \emph{seen-} and \emph{unseen-transformed} categories (see Appendix~\ref{app:additional}), which show that transformed images from \emph{unseen-transformed} categories are mainly confused with \emph{seen-transformed} categories.

\noindent {\bf Color, Convolutional and Noise Transformations lead to Higher Across-Category Robustness than Geometric Transformations.}
The increase of the across-category accuracy with the number of \emph{seen-transformed} categories is common among  transformations. Yet, Fig.~\ref{fig:fig2}b and c shows that the rate at which the across-category varies is different depending on the transformation. Observe that for color transformations, just few \emph{seen-transformed} categories are sufficient  to achieve a high across-category accuracy. This may be because the network is already quite robust to color transformations even training only with non-transformed images, as shown in Fig.~\ref{fig:fig2}a. For convolutional transformations, the increase of the across-category accuracy is not as high as for color transformations, but it is remarkable. The across-category accuracy of the noise transformations is in-par with the convolutional in ImageNet but not in FaceScrub. Note that the within-category accuracy for noise transformations in FaceScrub  is between $10\%$ to $20\%$ lower than for other transformations. Thus, in FaceScrub the amount of noise impairs both within- and across-category robustness, which suggests that the low across-category accuracy is not a characteristic of the noise transformation but of the overall accuracy in FaceScrub.

For geometric transformations, the across-category accuracy is lower than the rest of transformations, specially for rotations (the increase is only of a few percent when all categories are seen transformed). This may be because spatial re-arrangement of the object are involved in geometric transformations but not in the rest. These re-arrangements are larger for rotations of $90$ or $180$ degrees than for thinning, and rotations have much lower across-category accuracy than thinning. Also, to achieve across-category robustness for spatial   re-arrangements requires capturing the long-range dependencies between object parts, and this has been shown to be difficult for feed-forward architectures~(\cite{minimal}, \cite{engstrom2017exploring}, \cite{azulay2019deep}, \cite{jaderberg2015spatial}).

\noindent {\bf Across-Category Robustness may depend on the Visual Homogeneity of the Object Categories.}
For convolutional and noise transformations, the across-category generalization in FaceScrub increases linearly with respect to the number of categories seen transformed, while in ImageNet it increases much more slowly (note that in ImageNet the number of seen transformed categories is in logarithmic scale). These difference between datasets may be because the object categories in FaceScrub are more visually homogeneous than in ImageNet,~\ie FaceScrub {\color{black} share common configurations across different facial identities while ImageNet is composed of a varied set of more distinct object categories. Thus, the visual homogeneity of the categories may facilitate  across-category generalization (assuming that the invariant features can be more effectively transferred across categories that share comparable visual structures).}

\subsection{Results on Invariance to Transformations at the  Individual Neural Level}

\begin{figure*}[t!]
\centering
\scriptsize
\includegraphics[width=\textwidth]{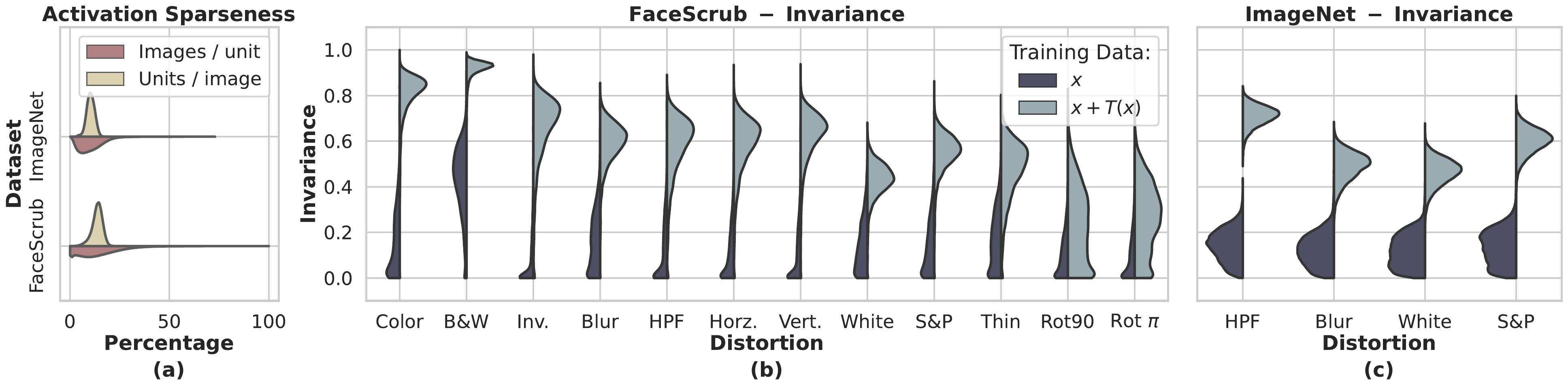}
\caption{\emph{Active Neurons and Invariance when all categories are \emph{seen-transformed}.} (a) Distribution of the number of active neurons per image and the number of images that activate a neuron. (b) and (c) Violin plots of the invariance coefficient among neurons in the penultimate layer, for FaceScrub and ImageNet, respectively. The invariance to each transformation is shown for  networks trained with the transformation applied to all categories (indicated with $x + T(x)$) and for networks trained only with non-transformed images ($x$). }
\label{fig:fig3}
\vspace{1cm}
\end{figure*}

\begin{figure*}[t!]
\centering
\scriptsize
\includegraphics[width=\textwidth]{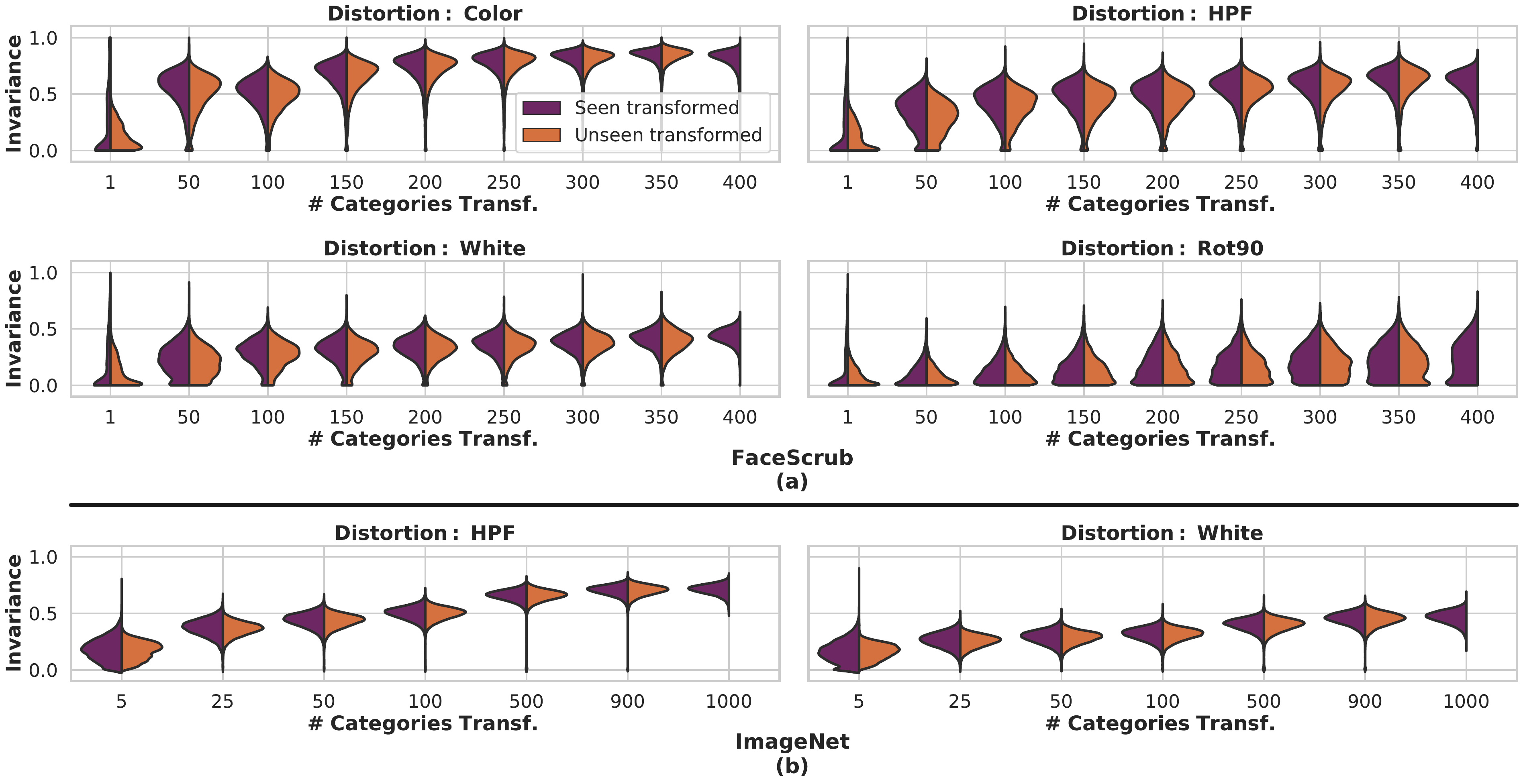}
\caption{\emph{Invariance for Different Number of  \emph{Seen-transformed} Categories}. (a) and (b) Violin plots of the invariance coefficient among  neurons at the penultimate layer for FaceScrub and ImageNet, respectively. Different number of  \emph{seen-transformed} categories are displayed and the invariance coefficient is reported separately for  \emph{seen-} and \emph{unseen-transformed} categories.}
\label{fig:fig4}
\vspace{1cm}
\end{figure*}

Next, we evaluate the amount of invariance using the measure introduced in section~\ref{sec:invariance}.

\noindent {\bf Less than $20\%$ of the Images Activate a Neuron.}
Recall that the threshold that determines if a neuron is active, $\tau$, is  the compromise between two factors: \emph{(i)} neurons are only active for few images because they are tuned to specific patterns,~\ie $\tau$ is as high as possible, \emph{(ii)} only the activity in active neurons matters for robustness,~\ie ablating all the inactive neurons in an image should not drop the accuracy whereas ablating all the active neurons in an image should bring the accuracy to chance. In practice, we implement a grid search to determine $\tau$ by
evaluating values between $0$ to $1$ in steps of $0.05$. We select the highest $\tau$ that leads to a drop of $1\%$ of the accuracy when ablating the inactive neurons.  This results in $\tau=0.1$ for FaceScrub and $\tau=0.05$ for ImageNet. We validate that neurons are active only for a small subset of images,~\ie the cardinality of $\mathcal{A}_k$ is small for most neurons $k$, and also that each image activates a subset of neurons,~\ie the cardinality of $\{k | x \in \mathcal{A}_k\}$ is small for most images $x$. 
In Fig.~\ref{fig:fig3}a we show the distributions of both cardinalities using violin plots. Notably, between $1$ to $20\%$ of the images activate a neuron, and between $10$ to $20\%$ of the neurons are active in an image. Thus, neurons are tuned to patterns that appear in few images and in most images only a subset of neurons is active, which fulfills factor \emph{(i)}.
In Appendix~\ref{app:additional}, we report the network's accuracy obtained from ablating active and inactive neurons. We observe that when the inactive neurons are ablated ($80-90\%$ of the neurons per image), the network's accuracy  decreases less than $1\%$ overall. When the active neurons are ablated ($10-20\%$ of the neurons per image), the network's accuracy is very low for all transformations. This low but above chance accuracy suggests that the neural activity of the inactive neurons may play a minor but redundant role towards the network's robustness, which approximately fulfills factor \emph{(ii)}.

{\color{black}
In Appendix~\ref{app:additional}, we confirm that \cite{goodfellow2009measuring} proposal to measure invariance only in the top-$1\%$ of the neural activity per neuron is incomplete. The results in the appendix provide evidence that supports the notion that a substantial proportion of accuracy performance is still attributed to neurons that were excluded by the threshold of the top-$1\%$ proposed by \cite{goodfellow2009measuring}. Specifically, applying the top-$20\%$ threshold allows to eliminate a large number of neurons that significantly contribute to the network's accuracy. This observation justifies using the procedure that we introduced as it is a much more accurate procedure to filter out neurons that are inactive most of the time. }

%by comparing Fig~\ref{fig:fig3}a and b with Fig.~\ref{fig:fig2}a

\noindent {\bf Invariance Increases when the Number of \emph{Seen-transformed} Categories Increases.}
Recall that the invariance coefficient, $I_k$, is evaluated when the neurons are active. In Fig.~\ref{fig:fig3}b and c, we show the distribution of $I_k$ among neurons in the penultimate layer  for networks that are trained with or without transformations. All transformations lead to more invariance than when the training is only with non-transformed images. In Appendix~\ref{app:additional}, we show that this invariance builds up across layers, which is in accordance with previous works~(\cite{goodfellow2009measuring}, \cite{poggio2016visual}).

Next, we aim to explore the relationship between invariance and the across-category robustness. Fig.~\ref{fig:fig4} shows the distribution of $I_k$  among neurons in the penultimate layer when the number of \emph{seen-transformed} categories  is increased. The distribution for the  \emph{seen-} and \emph{unseen-transformed}  categories is shown separately (see Appendix~\ref{app:additional} for all transformations). We observe that the amount of invariance remarkably increases  when the number of  \emph{seen-transformed} categories is increased. Also,  the amount of invariance  for  the  \emph{seen-} and \emph{unseen-transformed} categories   is nearly identical (except when the number of  \emph{seen-transformed} categories is low because there is almost no invariance). Thus, invariance in the network is not category-dependent and emerges alongside  across-category robustness. This is strong evidence that invariance contributes to robustness when the number of \emph{seen-transformed} categories is large.  

{\color{black}
It is worth noting that while most neurons develop invariance to transformations through training, it is unclear what are the neural mechanisms that emerge when there is no invairance. In the case that invariance does not emerge, networks are not robust to transformations of \emph{unseen-transformed} categories, but networks are still robust to transforimation of the \emph{seen-transformed} categories. A plausible hypothesis to explain this phenomenon is that some neurons may exhibit specialization towards the transformations such that the transformations can be handled for \emph{seen-transformed} categories but not for \emph{unseen-transformed} categories. In order to verify the presence of neurons specialized to the transformation, we modified Equation~\eqref{eq:H} to calculate a specialization score, which yields a value of $1$ if a neuron is selective for $T(x)$ but not $x$, or $-1$ if a neuron is selective for $x$ but not $T(x)$. In Appendix~\ref{app:additional}, our results demonstrate that for the transformations that when included in the training set do not lead to the emergence of invariant representations, there are neurons specialized to $x$ and other neurons to $T(x)$, but there is also a range of neurons in between. This can be observed in Fig.~\ref{fig:specialization} (Appendix~\ref{app:additional}), specially for the case of rotations, as the violin plot is spread out across different degrees of specialization. In conclusion, the neural mechanisms when there is no emergent invariance are related to specialization to a certain degree, and more research is needed to fully understand them. }

\begin{figure*}[t!]
\centering
\scriptsize
\includegraphics[width=\textwidth]{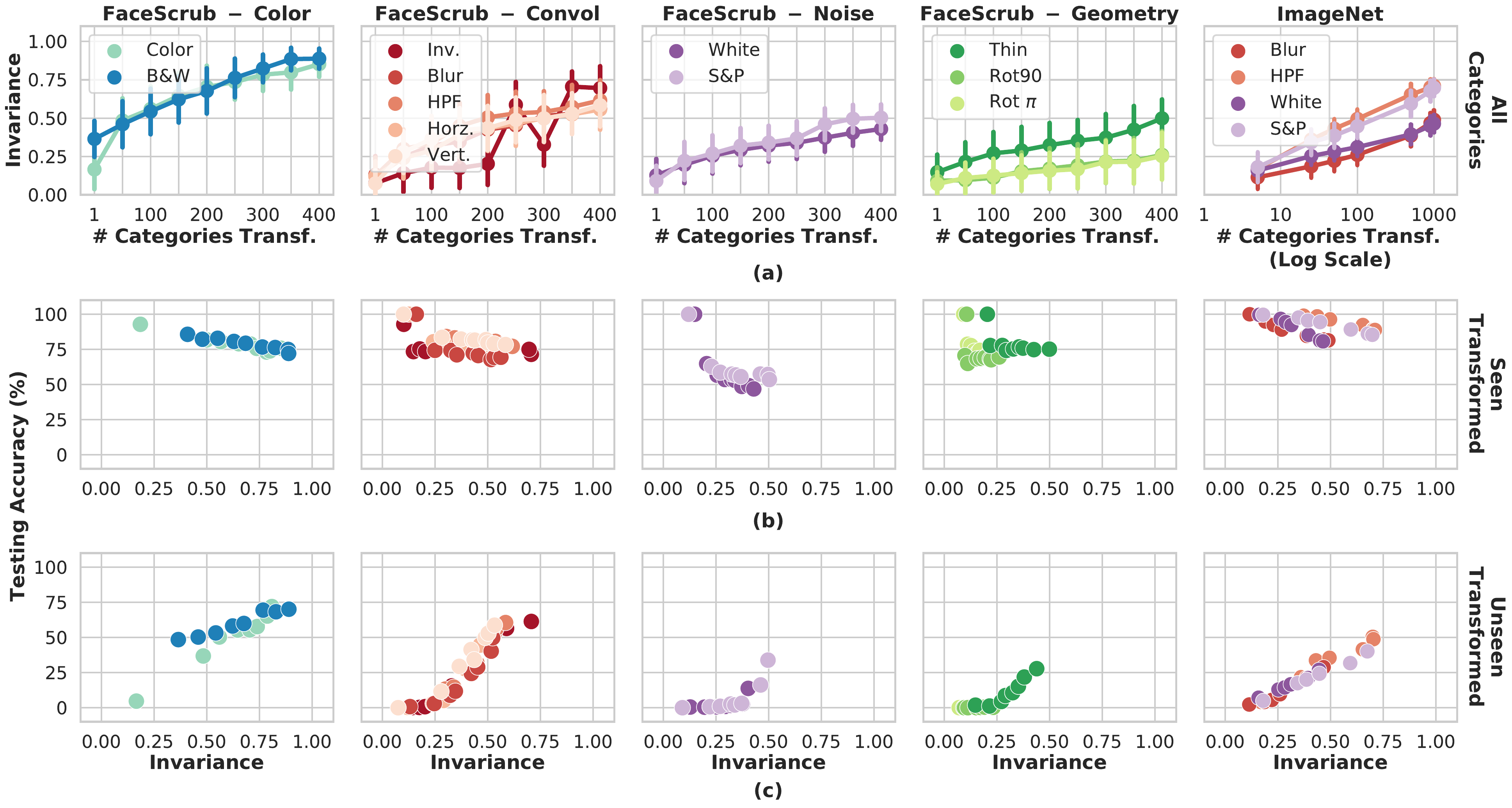}
\caption{\emph{Invariance and Robustness.} (a) Amount of invariance for different number of  \emph{seen-transformed} categories, transformations and datasets. (b) and (c) Relationship between amount of invariance and within- and across-category accuracy, respectively.   Each dot represents a network trained with different number of  \emph{seen-transformed} categories.}
\label{fig:fig5}
\vspace{1cm}
\end{figure*}

\noindent {\bf Invariance Explains the Across-Category Robustness.} Next, we further analyze the relationship between invariance and across-category robustness by scrutinizing the differences among transformations.
Fig.~\ref{fig:fig5}a shows the amount of invariance for all transformations and datasets as the number of  \emph{seen-transformed} categories is increased. We observe that the trends of invariance follow that of across-category robustness. To quantify whether there is a direct relationship between them,  in Fig.~\ref{fig:fig5}b and c we depict the invariance and robustness for each network trained with a different number of  \emph{seen-transformed} categories. Observe that invariance is not needed to achieve within-category robustness (Fig.~\ref{fig:fig5}b), whereas the extent of across-category robustness is  correlated with the amount of invariance for all transformations (Fig.~\ref{fig:fig5}c).  Thus, the network  develops properties different from invariance to achieve robustness, but as we increase the number of  \emph{seen-transformed} categories the amount of invariance increases.

{\color{black}
\noindent {\bf Results are consistent across network architectures.}
The achievement of invariance can be affected by the network architecture and capacity chosen for the task, presenting a potential point of concern. Although we attempted to address this concern by experimenting with AlexNet and ResNet-18, the differences in the datasets on which these models were trained leave open the question of whether the selection of different networks could produce distinct patterns. To mitigate this concern, we employed the Inception model \citep{szegedy2015going} and conducted the same experiment on FaceScrub. Our analysis demonstrates a consistent pattern wherein invariance gradually emerges as the number of transformation-seen categories increases (see Appendix~\ref{app:additional}). This finding again suggests that the relationship between invariance and the number of transformation-seen categories during training is a generalizable phenomenon, irrespective of the specific network architecture employed.
However, it is intriguing to note that Inception exhibits a higher number of transformation-invariant neurons than AlexNet, indicating that network architecture and capacity may influence the attainment of invariance. Despite this, the fundamental observation that invariance arises at the neural level and is closely linked to robust performance remains valid. Future studies exploring the association between network capacity and invariance will be of great interest. }

\vspace{-0.1cm}
\section{Discussions \& Conclusions}

We have demonstrated that increasing the number of  \emph{seen-transformed} categories results in an increase of two network properties: the amount of invariance to transformations and the robustness to transformations of \emph{unseen-transformed} categories. This suggests that invariance drives robustness to transformations depending on the number of \emph{seen-transformed} categories. Furthermore, we have shown that invariance emerges at the individual neural level, which adds to the growing body of literature that uses individual neurons as the elemental building blocks to analyze DCNNs~(\cite{zeiler2014visualizing}, \cite{zhou2018revisiting}, \cite{olah2018building}). Key open questions derived from our results that will be tackled in future works are understanding why invariant representations emerge for some transformations, and  what are the neural mechanisms for robustness when there is a lack of invariance (\eg for geometric transformations and also for small number of \emph{seen-transformed} categories).

{\color{black}
One potential limitation of our study is that it examines how networks attain invariance for individual types of transformations independently, whereas multiple types of transformations often co-occur in real-world scenarios. It is worth exploring how individual neurons develop invariance in the presence of multiple transformation types. We anticipate that neurons would form clusters based on relevant transformations and demonstrate distinct patterns across different levels of visual processing. As this exploration would necessitate additional experimental complexity, we concede that it remains an open question that needs to be addressed in future research.

Another promising avenue for future research is to explore how individual neurons respond to transformations that fall outside the range of the training transformations. While our training approach focused on generalizing invariance across different object categories, it remains an open question whether our findings hold under such conditions. The literature suggests that neural network models face significant challenges when expanding their knowledge to unseen conditions beyond the training regime~(\cite{montero2021role}, \cite{Schott2021VisualRL}). We encourage further investigation of this topic in future research.

Furthermore, in building bridges between machine learning and neuroscience, the current work adds to the existing body of literature that seeks to understand how visual experience drives the development of robust recognition systems (\cite{le2001early}, \cite{wood2018development}, \cite{jang2021noise}). More specifically, our results suggest that robust recognition behavior across categories can be achieved as individual neurons become tuned to be invariant to experienced transformations. This enriches our understanding of how biological visual systems may develop invariant representations of objects while maintaining their sensitivity to object-specific features, one of the central questions in computational neuroscience (\cite{palmeri2004visual}, \cite{peissig2007visual}, \cite{tsao2008mechanisms}, \cite{dicarlo2012does}, \cite{tacchetti2018invariant}). Our computational findings also agree with previous experimental observations made in studies of single neurons. For example, the prefrontal cortex neurons have been found to become invariant to noise after learning to recognize noisy images \citep{rainer2000effects}. Additionally, our training paradigm may account for the observation that a large number of cells in the macaque showed viewpoint invariance to objects that the monkey had never seen before \citep{bao2020map}.

The current study motivates future endeavors in determining other contributing factors to invariance. We found that the network's ability to learn invariance was significantly lower when geometric transformations were evaluated. One potential explanation is that our training scheme was only tested within a class of feedforward networks, while the brain heavily relies on recurrent computations (\cite{felleman1991distributed}, \cite{lamme2000distinct}). Recurrent processing has been shown to confer a critical advantage in performing robust object recognition (\cite{wyatte2012limits}, \cite{o2013recurrent}, \cite{spoerer2017recurrent}, \cite{kar2019evidence}). It has been also suggested that recurrent processing is actively engaged when human observers perform mental rotation (\cite{shepard1971mental}, \cite{roelfsema2016early}), which may give a clue to explain why the tested networks in the current study failed to achieve invariance to geometric transformations. Therefore, it will be of considerable interest for future studies to investigate the efficacy of recurrent processing in acquiring invariance to various types of transformation. }

\section*{Acknowledgements}
We are grateful to Tomaso Poggio for his insightful advice and warm
encouragement. This work is supported by the Center for Brains, Minds and Machines (funded by NSF STC award
CCF-1231216), Fujitsu Laboratories Ltd. (Contract No. 40008401 and 40008819), the MIT-Sensetime Alliance
on Artificial Intelligence and the R01EY020517 grant from the National Eye Institute (NIH).

{\color{black}
\section*{Code availability} 
Code to reproduce the results available in the following public github repository: \url{https://github.com/xboix/InvarianceRobustness} }

%\section*{Statement of Broader Impact}
%This work tackles the long studied but still obscure area of generalization in DCNNs on two fronts. First, it analyzes the limits of robustness to transformations by examining the extensibility of robustness to categories that are not seen transformed during training. Our results reveal weaknesses of the generalization capability of DNNs as  robustness may degrade for categories that are not seen transformed during training. This phenomenon is severe for geometric transformations such as rotations. This could potentially be used to trick DCNNs in real-life applications. Nevertheless, understanding such weaknesses is necessary to build better and more robust DCNNs. 

%Secondly, by studying the role of invariant representations in individual neurons, this work identifies the important relationship between invariance and robustness for categories not seen transformed during training. By demonstrating the significance of invariant representations in driving robustness, this paper advocates for training methods that enforce invariance in the hidden layers. 
%The understanding of generalization in DCNNs is a critical area that will provide  strides in the advancements of modern computer vision and bring it closer to or beyond its rivaling human vision. 

%A negative consequence of this research is that it has unveiled weaknesses of DCNNs to certain transformations, which could potentially be used to trick DCNNs currently in practice. Nevertheless, understanding such weaknesses is necessary to build better and more robust DCNNs. 

\small

\printbibliography
%\bibliography{biblio.bib}
%\bibliographystyle{abbrv}

\clearpage
\appendix
%\section{Methods}

%In designing the experimental paradigm for FaceScrub, a total of 117 experiments were conducted for twelve transformations: ‘color rotation’, ‘grayscale’, ‘contrast inversion’, ‘blurring’, ‘high pass filtering’, ‘horizontal filtering’, ‘vertical filtering’, ‘white Gaussian noise addition’, ‘salt and pepper noise addition’, ‘thinning’, ‘90$^o$ rotation’, ‘180$^o$ rotation’. For each transformation, nine networks were trained such that the number of categories transformed during training, henceforth “cutoff”, were 1, 50, 100, 150, 200, 250, 300, 350, and 388, respectively. Nine experiments involved training a DNN on untransformed images corresponding to “none” transformation at each cutoff. 

%In the case of ImageNet, the values for cutoff were 5, 25, 50, 100, 500, 900, and 1000 respectively for ‘blurring’, ‘high-pass filtering’, ‘white noise’, and ‘salt and pepper noise’ transformations. A single network was trained on untransformed ImageNet, which was reused in plots for each cutoff value. 

%For all experiments, the test-set performance was measured separately for the categories which were transformed during training and the categories which remained untransformed. The performances are termed ‘within-category’ and ‘across-category’ generalization, respectively, throughout the text. In FaceScrub, we report the performance as the top-1 accuracy, whereas, for ImageNet top-5 accuracy is reported.

\section{Transformations}
\label{app:transformations}
The transformations were applied to the dataset as an input layer in the GPU, such that the transformations were computed at runtime with every forward pass.

For color transformation, the hue was rotated 180$^o$ in the hue space. The grayscale operation was performed using built-in TensorFlow functions.

For blur transformation, a Gaussian filter was convolved with the input image. The standard deviation of the Gaussian was selected such that the top-1 accuracy of a network trained on the untransformed FaceScrub dataset was less than $10\%$. To do so, we increased the standard deviation in steps of $0.5$ until the accuracy was less than $10\%$, and then we stopped. The horizontal and vertical filters have a filter size of $3 \times 3$ of form $[-1, 0, 1]$. The high-pass filter has a filter size of $5 \times 5$ of form $[-1, 2, 4, 2, -1]$ in both directions.

Noise based transformations were applied by adding white noise and salt and pepper noise to the original images. The standard deviation of noise was selected with the same procedure as the blur transformation. For the white noise, the standard deviation was increase in steps of $25$. The salt and pepper noise randomly sets half of the pixels to either $0$ or $255$.

Geometrical transformations rotated the input images anti-clockwise with angles 90$^o$ and 180$^o$, respectively. For thinning, the image width was reduced to its half and padded with zeros on both sides.

In ImageNet, the transformations were applied with parameters such that the top-1 accuracy was under $15\%$, using the aforementioned procedure to set the parameters of the transformations.

\section{Datasets and Networks}
\label{app:dataset}
\noindent {\bf Datasets.} We use two datasets: FaceScrub~\citep{ng2014data} and ImageNet~\citep{deng2009imagenet}. FaceScrub is a face recognition dataset containing over 50,000 images of 388 individuals. ImageNet is an object recognition dataset with around 1.4 million images of 1000 object categories.

When increasing the number of \emph{seen-transformed} categories, we use the same \emph{seen-transformed} categories before the increase and an additional set of categories randomly selected. The smallest number of  \emph{seen-transformed} categories are randomly selected. This procedure facilitates comparing results between different number of \emph{seen-transformed} categories, as it minimizes the dependency of the selected \emph{seen-transformed} categories.

\noindent {\bf Networks.} For FaceScrub, we used a modified AlexNet~\citep{krizhevsky2012imagenet} architecture with five convolutional layers followed by two fully connected layers. The modifications included removal of the max-pooling layers after the first two convolutional layers in order to preserve necessary resolution for face recognition. The rest of the network was exactly as AlexNet (\ie local response normalization, dropout, initialization parameters were the same as in AlexNet).  The networks were trained for a maximum of $45$ epochs (sufficient for convergence in all cases) with a learning rate of 1e-4 and a weight decay of 5e-4 for all transformations. These values were selected via grid search over 25 possible combinations of learning rates and weight decay, evaluated on $10\%$ of a held-out images of the training set (all transformations lead to the same hyperparameters). The optimization algorithm was stochastic gradient descent with a momentum of $0.9$ and a batch size of $32$.

For ImageNet, we trained Resnet18s~\citep{he2016deep} using the official tensorflow implementation and with the hyperparameters that come by default.

\noindent {\bf Preprocessing.} All input images were standardized before the first hidden layer. This did not change the accuracy of the network, and it was particularly important to facilitate robustness to transformations  without the need of adjusting the mean and standard deviation parameters of the batch normalization to the transformation. Before the standardization, the transformations were applied to the image, whose pixel values were between $0$ and $255$.

\section{Alternative Training Paradigms}
\label{app:alterparadigms}

A possible alternative to our paradigm could be considered based on training the network in two steps,~\ie  first training the \emph{seen-transformed} and then the \emph{unseen-transformed} categories. Yet, this  has limitations to analyze invariance because lack of across-category robustness in such two steps paradigm can be due to other reasons unrelated to invariance,~\ie the two steps paradigm introduces confounding factors. Namely, if the training of the \emph{unseen-transformed} categories is done by fine-tuning the network after training the network with the \emph{seen-transformed} categories, lack of robustness could be due to the so-called ``catastrophic forgetting''~\citep{kirkpatrick2017overcoming} of the transformation rather than lack of invariance. Otherwise, if the training of the \emph{unseen-transformed} categories is done without fine-tuning,~\eg by training a classifier on the representations learned for the \emph{seen-transformed} categories, lack of robustness could be because these representations may not be generalizing well to the \emph{unseen-transformed}, rather than lack of invariance. Training both sets at the same time, as in our proposed paradigm, controls these confounding factors.% and facilitates the analysis of the emergent mechanisms to achieve robustness to transformations.

Another potential alternative approach could be to utilize a meta-learning methodology (\cite{vinyals2016matching}, \cite{snell2017prototypical}), in which we can investigate the generalization of invariance learned from the training set to a novel set. Since networks are presumed to acquire the ability to learn more in the concept space, meta-learning could provide a beneficial environment to evaluate invariance less influenced by category bias and could potentially serve as an appropriate platform for testing generalization ability. The extent to which the results from our current paradigm align with those obtained from a meta-learning framework remains an open question.

\section{Additional Results}
\label{app:additional}
\begin{figure*}[t!]
\centering
\scriptsize
\includegraphics[width=\textwidth]{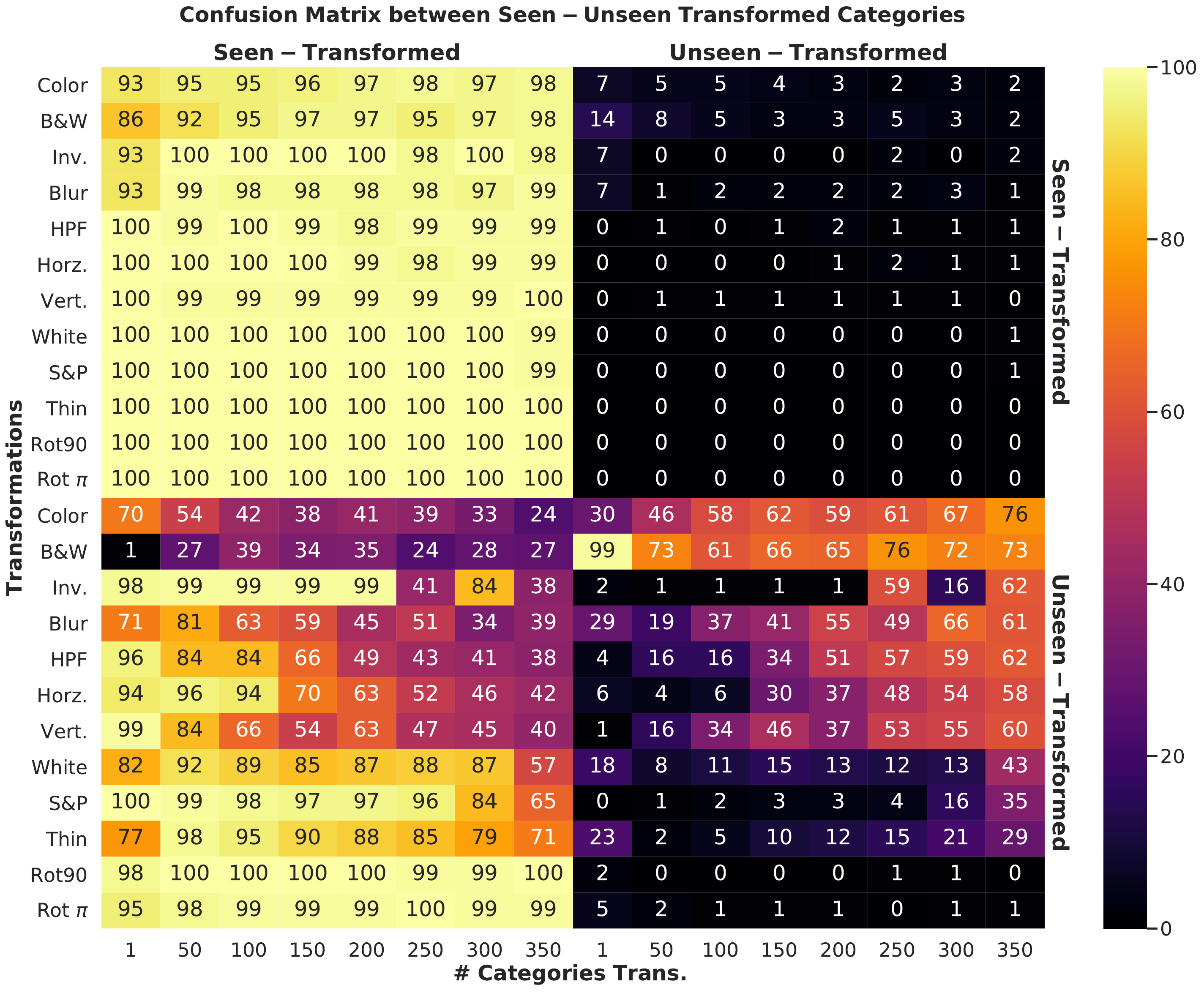}
\caption{\emph{Confusion Matrix Between Seen- and Unseen-Transformed Categories for FaceScrubs for all Transformations at different number of Categories transformed during training.}  Quadrant I: Percentage of Predictions for Unseen-Transformed Categories miss-classified as Seen-Transformed. Quadrant II: Percentage of Predictions for Seen-Transformed Categories correctly classified as Seen-Transformed. Quadrant III: Percentage of Predictions for Seen-Transformed Categories miss-classified as Unseen-Transformed. Quadrant IV: Percentage of Predictions for Unseen-Transformed Categories correctly classified as Unseen-Transformed.}
\label{fig:confusion}
\vspace{-0.2cm}
\end{figure*}

\noindent {\bf Confusion Matrices.} In Fig.~\ref{fig:confusion}, we show the confusion matrix between \emph{seen-} and \emph{unseen-transformed} categories. The table reports the percentage of images of one set of categories classified as the same or another set.  

\begin{figure*}[t!]
\centering
\scriptsize
\includegraphics[width=\textwidth]{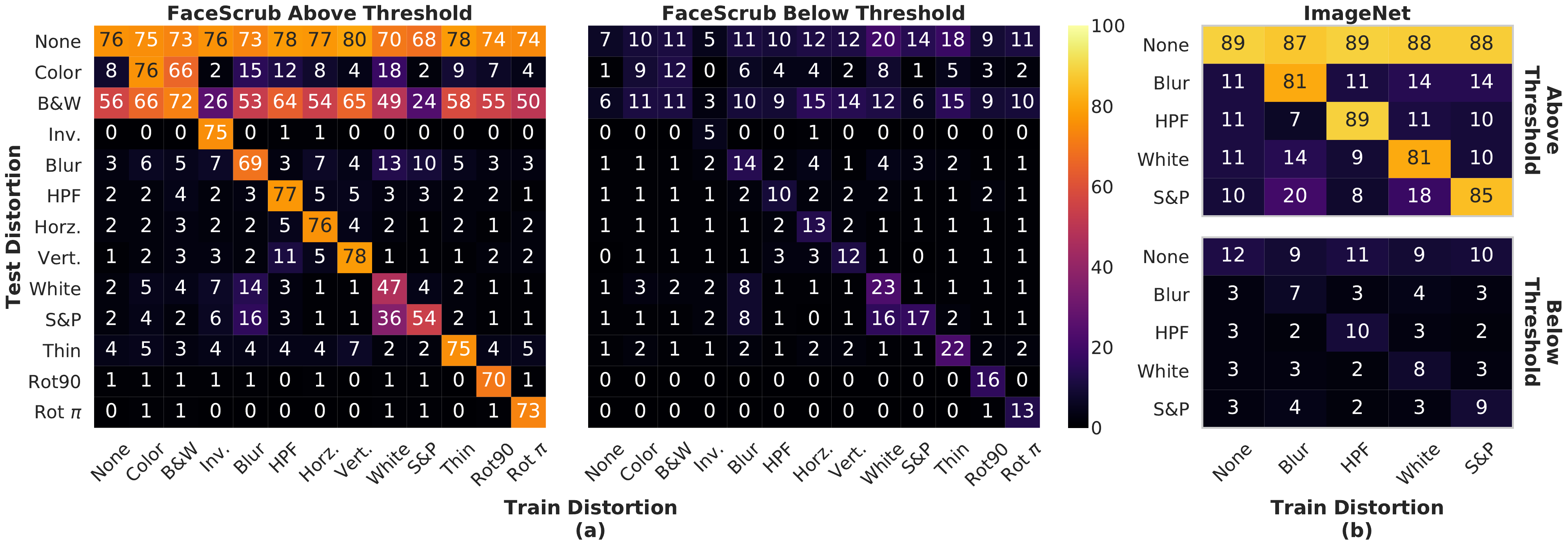}
\caption{\emph{Accuracy of Ablating Active and Inactive Neurons.} (a) and (b) Within-category accuracy after ablating inactive (left) and active (right) neurons for FaceScrub and ImageNet, respectively.  }
\label{fig:activeneurons}
\vspace{-0.2cm}
\end{figure*}

\noindent {\bf Active Neurons.} To validate that the activity of the active neurons,~\ie neurons with activity higher than $\tau$, capture all the activity relevant for the network's robustness, we performed two types of ablation experiments: set all activity below the threshold  to zero and set all the activity above the threshold to zero. The performance of the DCNN in these two conditions is shown in Fig.~\ref{fig:activeneurons}.

\begin{figure*}[t!]
\centering
\scriptsize

\includegraphics[width=0.5\textwidth]{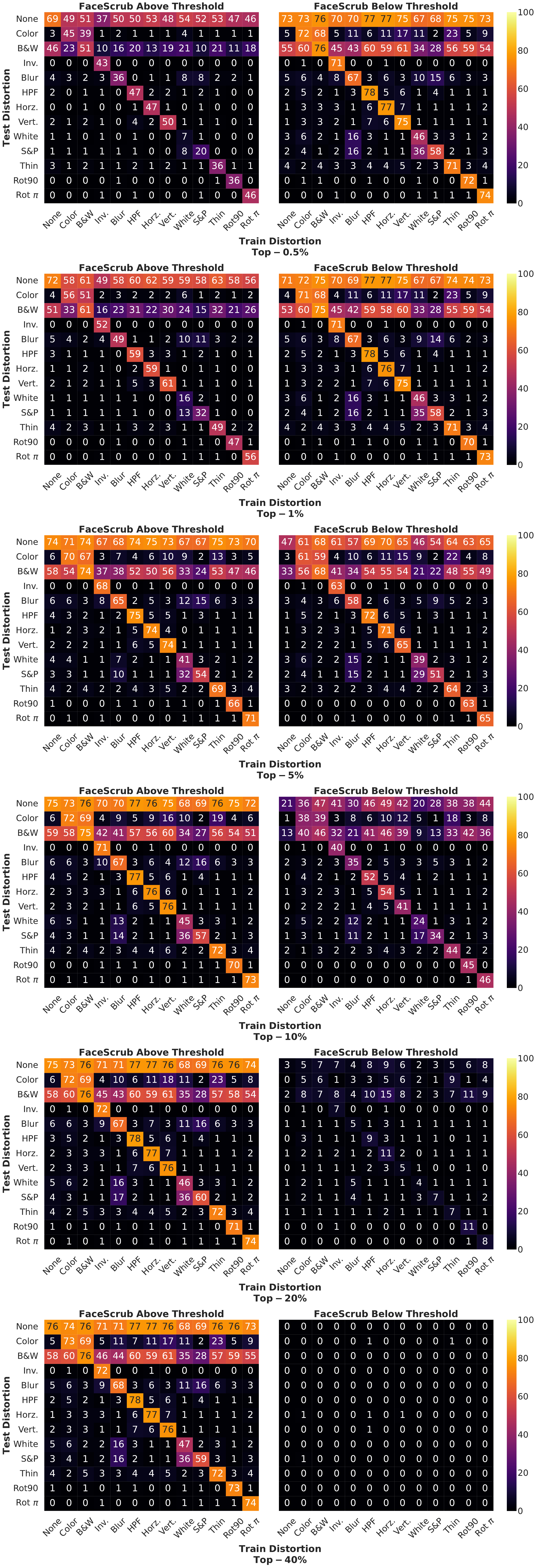}

\caption{\emph{Accuracy of Ablating Active and Inactive Neurons.} Within-category accuracy after ablating inactive (left) and active (right) neurons for FaceScrub by changing the threshold of the proposed method by~\cite{goodfellow2009measuring}. }

\label{fig:goodfellow}
\end{figure*}

\noindent {\bf Evaluation of the method by \cite{goodfellow2009measuring}.} To assess the effectiveness of \cite{goodfellow2009measuring}'s method, we varied the firing threshold from $0.5\%$ to $1\%$ (as  reported in the original study), $5\%$, $10\%$, $20\%$, and $40\%$, and measured the accuracies when ablating active and inactive neurons. The resulting accuracies are presented in Fig.~\ref{fig:goodfellow}.

\begin{figure*}[t!]
\centering
\scriptsize

\vspace{0.5cm}
\includegraphics[width=\textwidth]{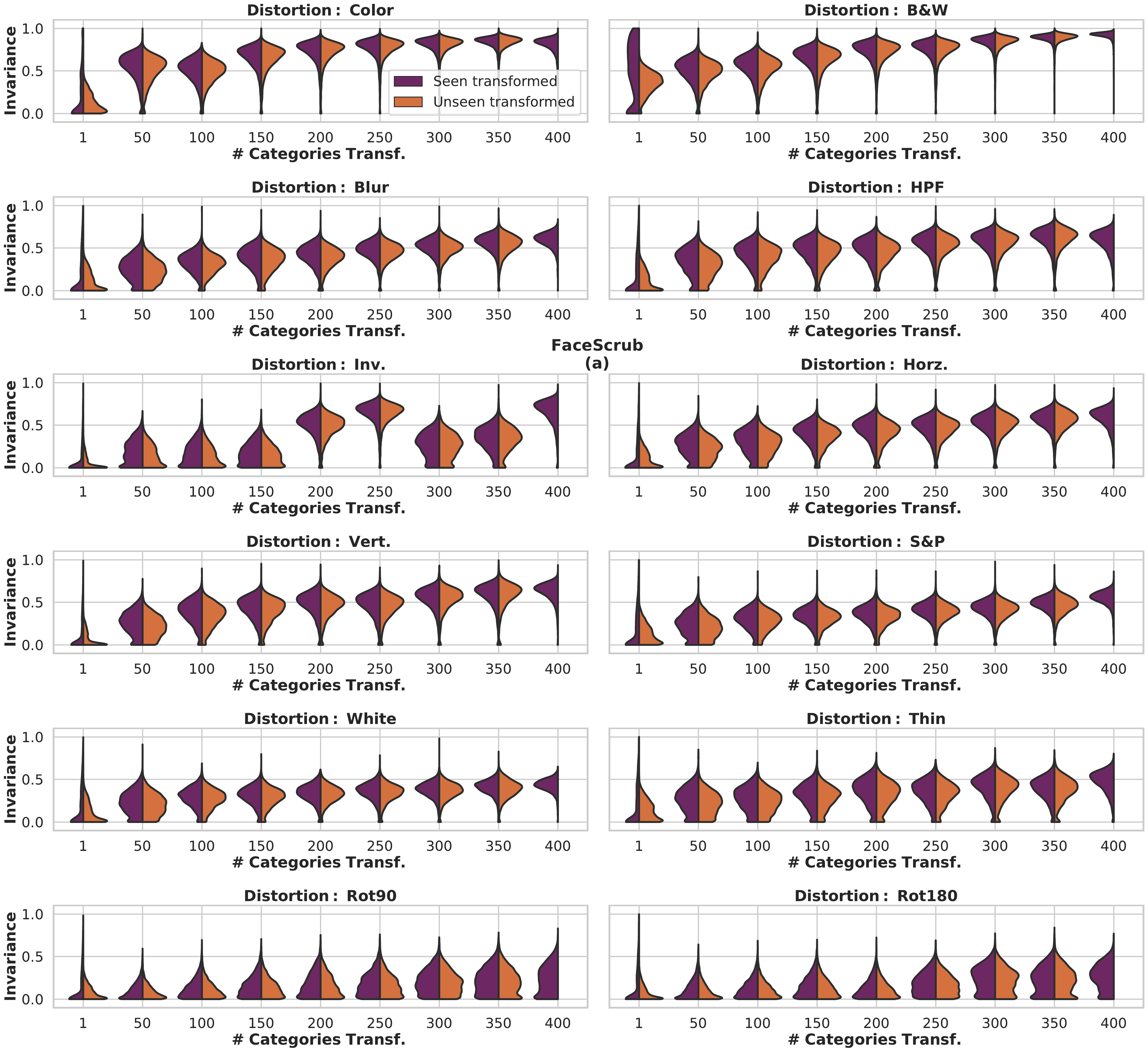}

% \begin{tabular}{@{}c@{}c@{}c@{}c@{}}
% \includegraphics[width=0.5\textwidth]{figs/paper_plots/Invariance_Distribution_color.pdf} &
% \includegraphics[width=0.5\textwidth]{figs/paper_plots/Invariance_Distribution_grayscale.pdf} \\
% \includegraphics[width=0.5\textwidth]{figs/paper_plots/Invariance_Distribution_low_pass.pdf} &
% \includegraphics[width=0.5\textwidth]{figs/paper_plots/Invariance_Distribution_high_pass.pdf} \\
% \includegraphics[width=0.5\textwidth]{figs/paper_plots/Invariance_Distribution_contrast.pdf} &
% \includegraphics[width=0.5\textwidth]{figs/paper_plots/Invariance_Distribution_horizontalfilter.pdf} \\
% \includegraphics[width=0.5\textwidth]{figs/paper_plots/Invariance_Distribution_verticalfilter.pdf} &
% \includegraphics[width=0.5\textwidth]{figs/paper_plots/Invariance_Distribution_salt_pepper.pdf} \\
% \includegraphics[width=0.5\textwidth]{figs/paper_plots/Invariance_Distribution_noise.pdf} &
% \includegraphics[width=0.5\textwidth]{figs/paper_plots/Invariance_Distribution_thinning.pdf} \\
% \includegraphics[width=0.5\textwidth]{figs/paper_plots/Invariance_Distribution_rotate90.pdf} &
% \includegraphics[width=0.5\textwidth]{figs/paper_plots/Invariance_Distribution_rotate180.pdf} \\
% \end{tabular}

\caption{\emph{Invariance for Different Number of  \emph{Seen-transformed} Categories}. Violin plots of the invariance coefficient among  neurons at the penultimate layer for FaceScrub. Different number of  \emph{seen-transformed} categories are displayed and the invariance coefficient is reported separately for  \emph{seen-} and \emph{unseen-transformed} categories.}
\label{fig:supp_faces_violin}
\end{figure*}

\begin{figure*}[t!]
\centering
\scriptsize
\includegraphics[width=\textwidth]{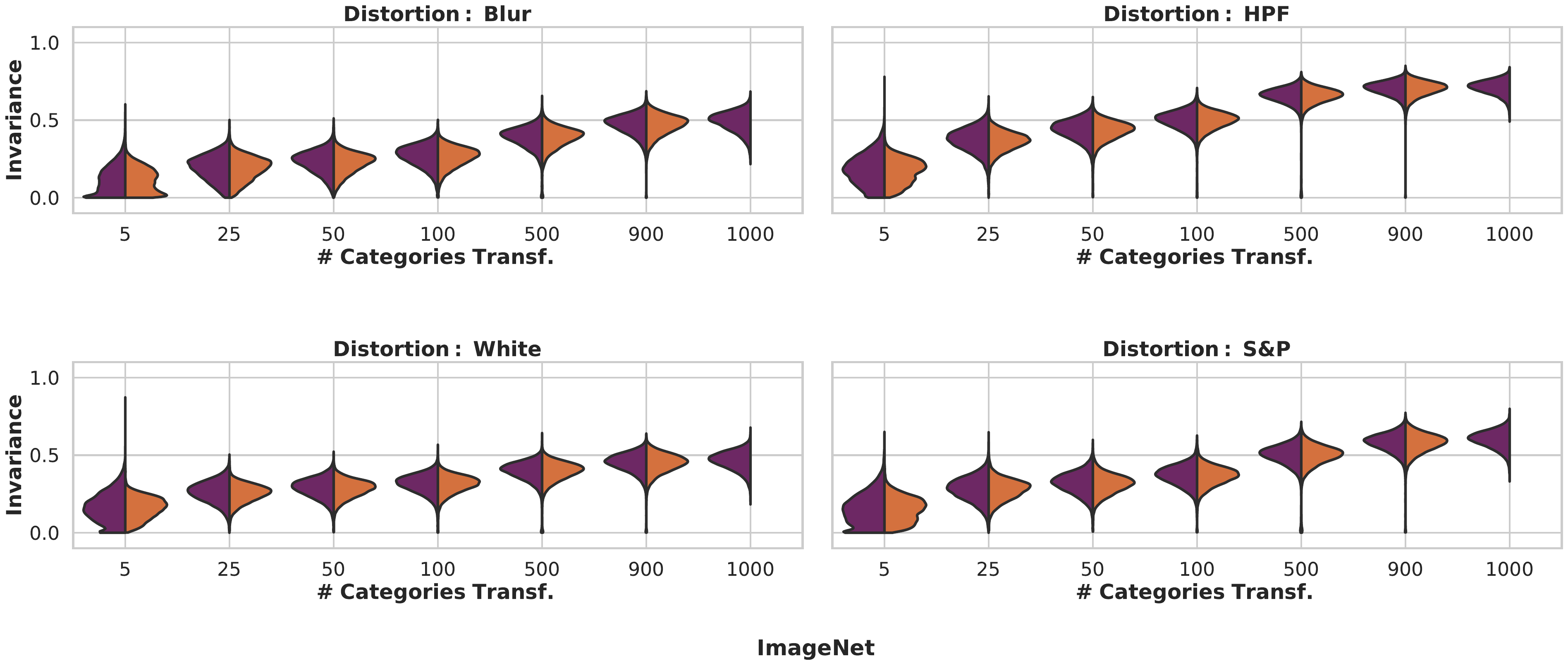}
\caption{\emph{Invariance for Different Number of  \emph{Seen-transformed} Categories}. Violin plots of the invariance coefficient among  neurons at the penultimate layer for ImageNet. Different number of  \emph{seen-transformed} categories are displayed and the invariance coefficient is reported separately for  \emph{seen-} and \emph{unseen-transformed} categories.}
\label{fig:supp_imagenet_violins}
\vspace{-0.2cm}
\end{figure*}

\begin{figure*}[t!]
\centering
\scriptsize
\includegraphics[width=\textwidth]{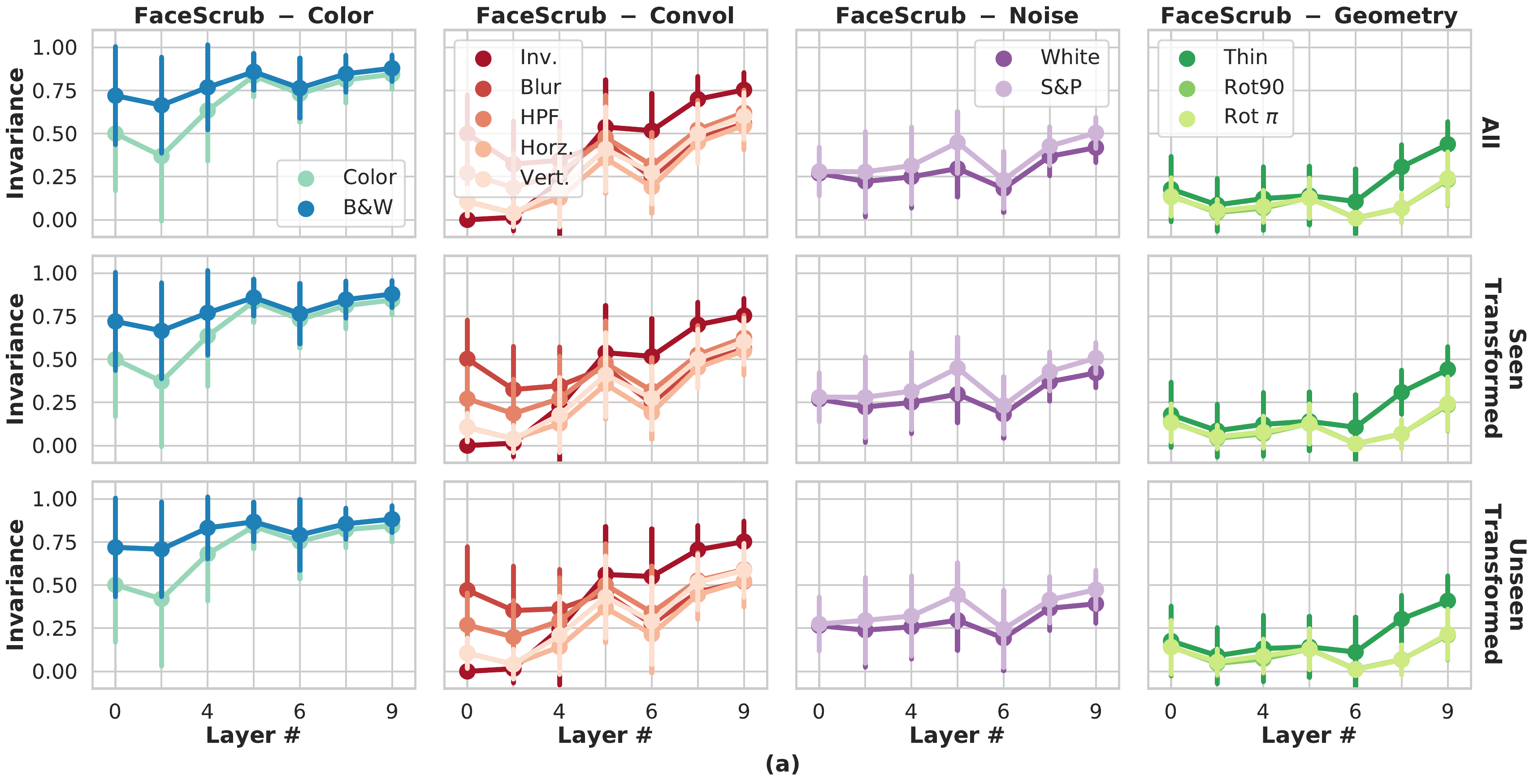}
\caption{\emph{Layer-wise Invariance.} Mean amount of invariance for different number of  \emph{seen-transformed} categories and transformations. Each row reports separately   \emph{seen-} and \emph{unseen-transformed} categories, or both.  }
\label{fig:layer_invariance}
\vspace{-0.2cm}
\end{figure*}

\noindent {\bf Invariance.} In Figs.~\ref{fig:supp_faces_violin} and ~\ref{fig:supp_imagenet_violins}, we display the invariance for the different transformations and number of categories \emph{seen-transformed} in FaceScrub and ImageNet, respectively.  Fig.~\ref{fig:layer_invariance} shows the amount of invariance at each layer.

\begin{figure*}[t!]
\centering
\scriptsize

\includegraphics[width=\textwidth]{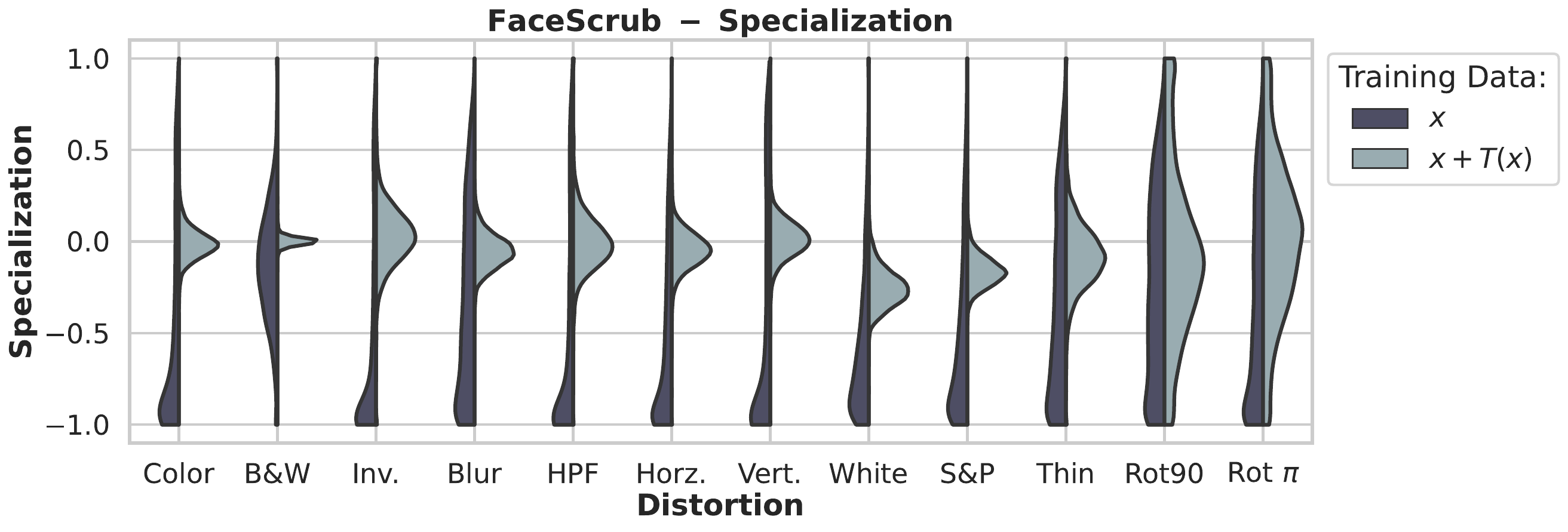}

\caption{\emph{Specialization in FaceScrub.} Violin plots of the specialization coefficient among neurons in the penultimate layer. The specialization to each transformation is shown for  networks trained with the transformation applied to all categories (indicated with $x + T(x)$) and for networks trained only with non-transformed images ($x$). }

\label{fig:specialization}
\vspace{-0.2cm}
\end{figure*}

\noindent {\bf Specialization.} To evaluate the selectivity of individual neurons towards transformed or non-transformed images, we modified Equation~\eqref{eq:H} to compute specialization scores as $\mathcal (f_k(T(x))  -  f_k(x)) / (f_k(T(x))  + f_k(x))$. The resulting score indicates the degree to which a neuron is selective towards either transformed or non-transformed images. A score of +1 indicates high selectivity for the transformed images, whereas a score of -1 indicates high selectivity for the non-transformed images. The specialization scores of individual neurons are shown in Fig.~\ref{fig:specialization}.

\begin{figure*}[t!]
\centering
\scriptsize

\begin{tabular}{@{}c@{}c@{}c@{}c@{}}
\includegraphics[width=1\textwidth]{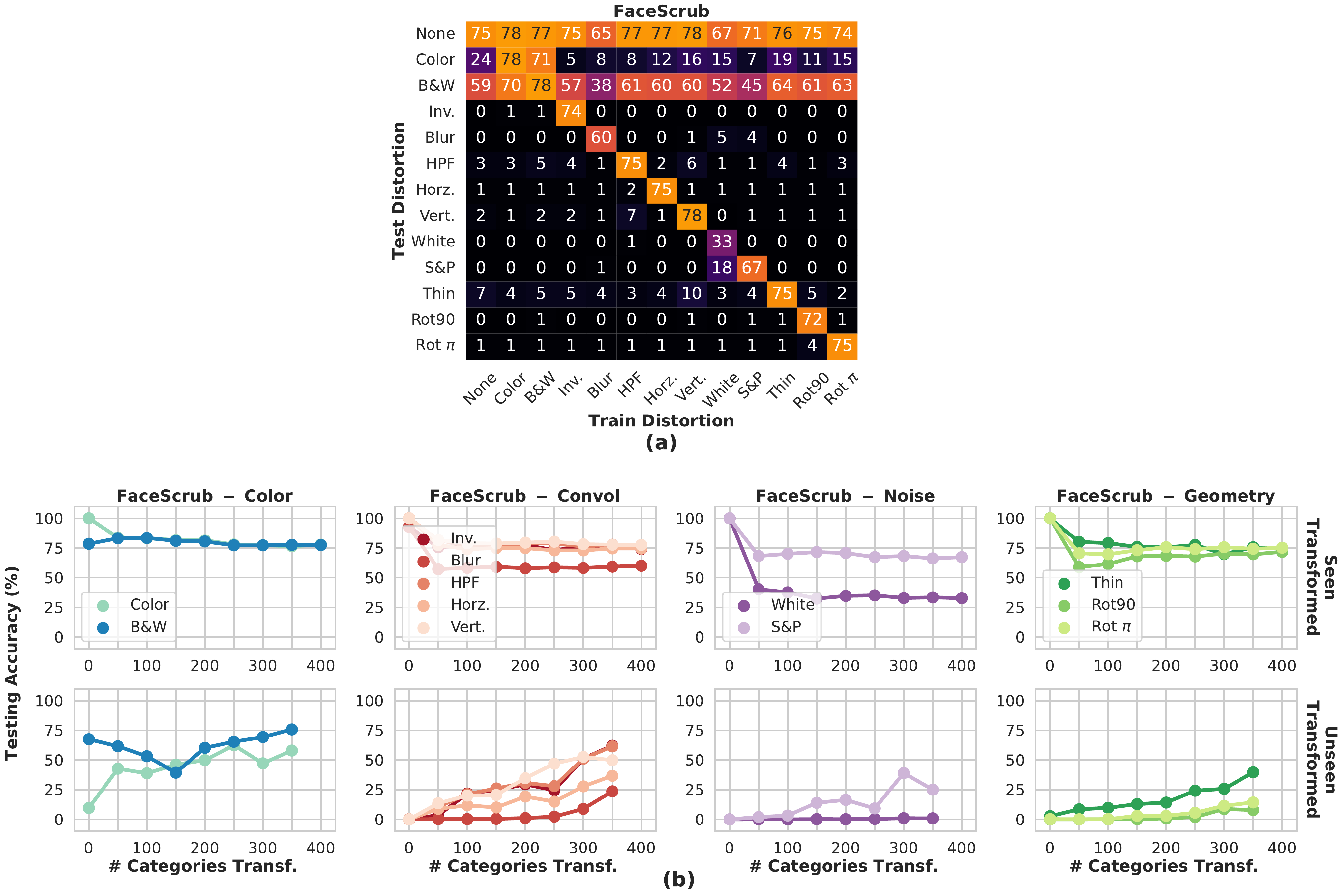}
\end{tabular}

% \caption{\emph{Within- and Across-Category Accuracy in FaceScrub using the Inception model.} (a) Within-category accuracy when all object categories are seen transformed. (b) Within-category accuracy (top) and accros-category accuracy (bottom), for different number of transformed categories in the training set. Each family of transformations is displayed separately. }

\caption{\emph{Within- and Across-Category Accuracy in FaceScrub using the Inception model.} Replication of Fig.~\ref{fig:fig2} by the Inception model. }

\label{fig:inception_accuracy}
\vspace{-0.2cm}
\end{figure*}

\begin{figure*}[t!]
\centering
\scriptsize

\begin{tabular}{@{}c@{}c@{}c@{}c@{}}
\includegraphics[width=1\textwidth]{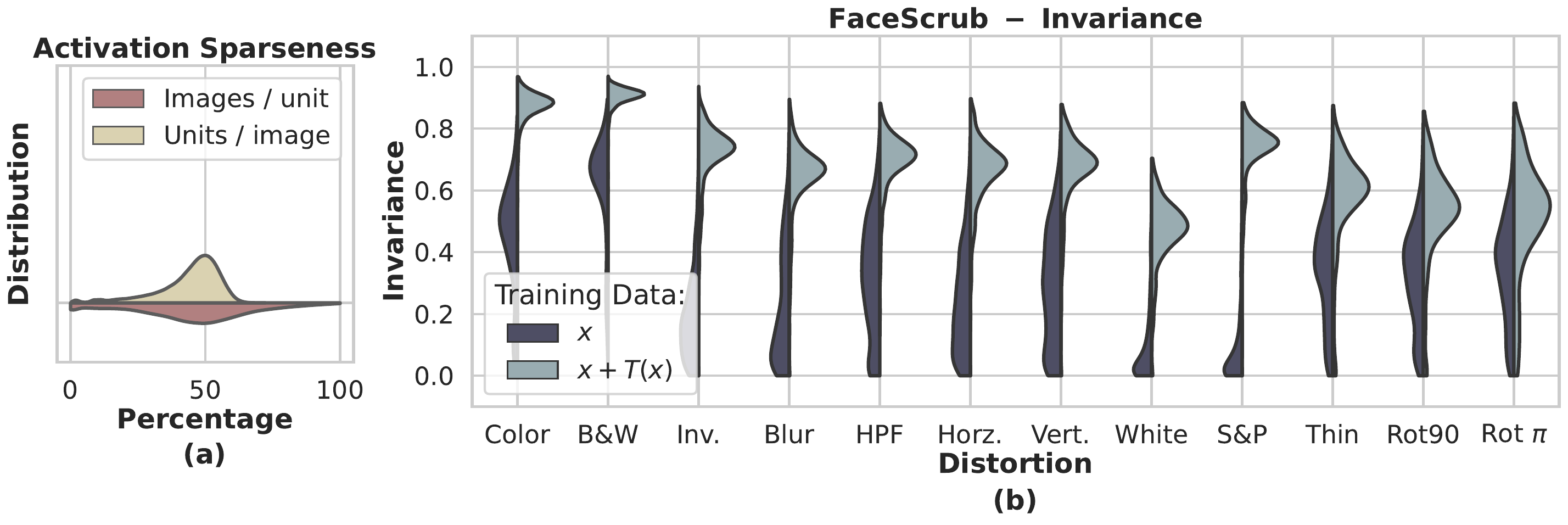} \\
\includegraphics[width=1\textwidth]{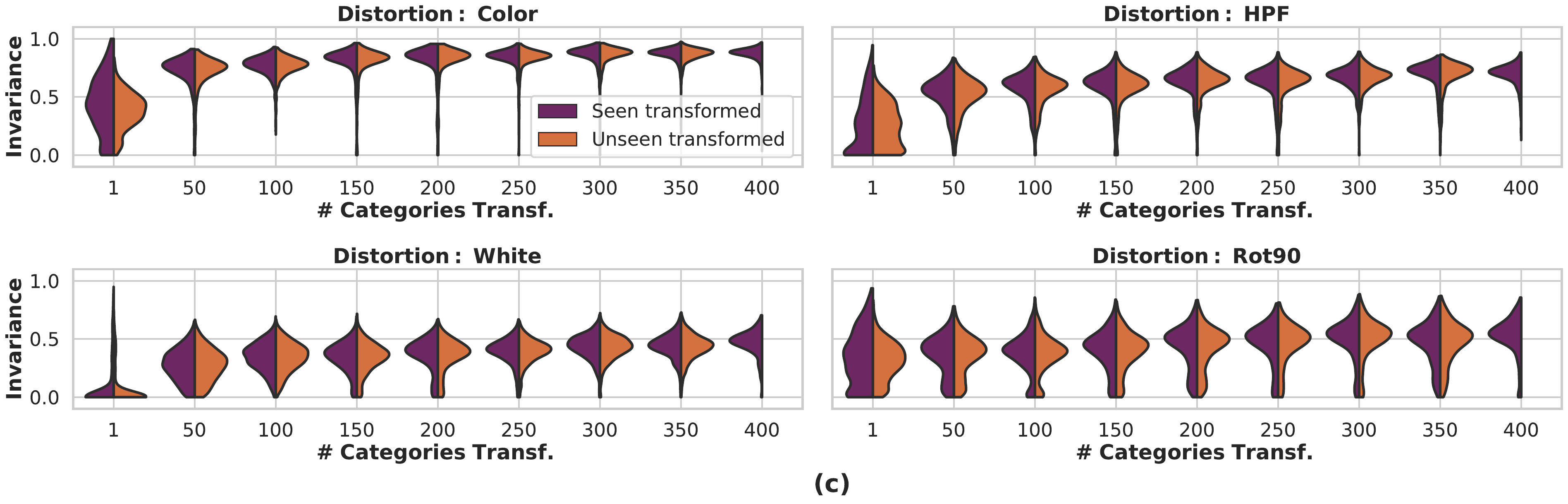} \\
\includegraphics[width=1\textwidth]{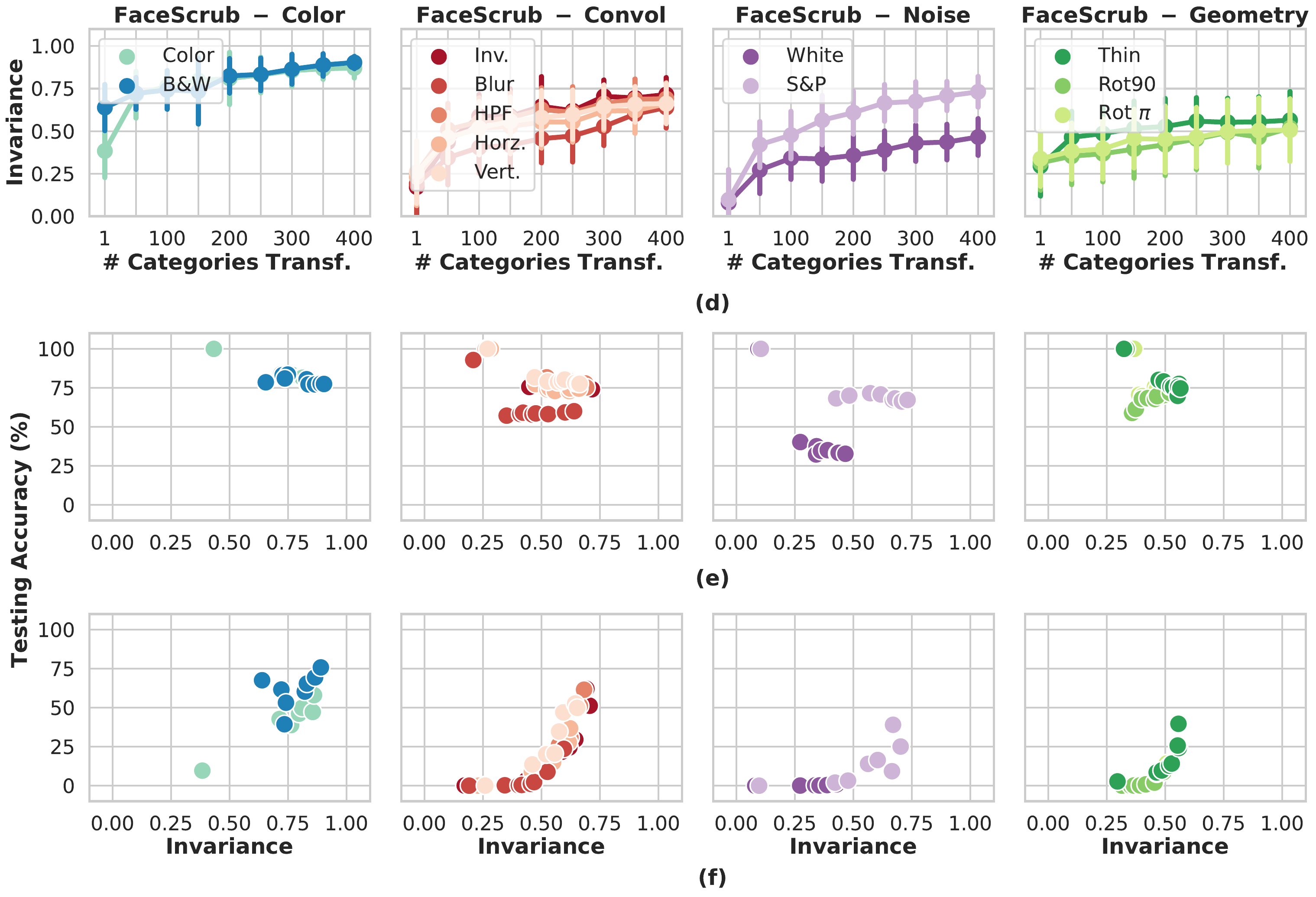} \\
\end{tabular}

\caption{\emph{Invariance coefficients of individual neurons in FaceScrub using the Inception model.} Replication of Figs.~\ref{fig:fig3}-\ref{fig:fig5} by the Inception model, shown in panels (a) through (f). }

\label{fig:inception_invariance}
\vspace{-0.2cm}
\end{figure*}

\noindent {\bf Inception.} In Figs.~\ref{fig:inception_accuracy} and ~\ref{fig:inception_invariance}, we replicate the previously reported findings of AlexNet for FaceScrub using the Inception model. We observe a consistent pattern of results that supports our initial assertion that invariance is achieved gradually as the number of transformation-seen categories increases and occurs at the level of individual neurons.

\clearpage

\end{document}